IAC-17-D3.3.2

# NETWORK OF NANO-LANDERS FOR IN-SITU CHARACTERIZATION OF ASTEROID IMPACT STUDIES

**Himangshu Kalita[a], Erik Asphaug[b], Stephen Schwartz[b], Jekan Thangavelautham[a]\***

[a] Space and Terrestrial Robotic Exploration Laboratory, Department of Aerospace and Mechanical Engineering, *University of Arizona, Tucson, Arizona 85721, United States,* jekan@arizona.edu
[b] Space and Terrestrial Robotic Exploration Laboratory, Department of Planetary Science and Lunar and Planetary Laboratory, *University of Arizona, Tucson, Arizona 85721, United States*

\* Corresponding Author

**Abstract**

Exploration of asteroids and comets can give insight into the origins of the solar system and can be instrumental in planetary defence and in-situ resource utilization (ISRU). Asteroids, due to their low gravity are a challenging target for surface exploration. Current missions envision performing touch-and-go operations over an asteroid surface. In this work, we analyse the feasibility of sending scores of nano-landers, each 1 kg in mass and volume of 1U, or 1000 $cm^3$. These landers would hop, roll and fly over the asteroid surface. The landers would include science instruments such as stereo cameras, hand-lens imagers and spectrometers to characterize rock composition. A network of nano-landers situated on the surface of an asteroid can provide unique and very detailed measurements of a spacecraft impacting onto an asteroid surface. A full-scale, artificial impact experiment onto an asteroid can help characterize its composition and geology and help in the development of asteroid deflection techniques intended for planetary defence. Scores of nano-landers could provide multiple complementary views of the impact, resultant seismic activity and trajectory of the ejecta. The nano-landers can analyse the pristine, unearthed regolith shielded from effects of UV and cosmic rays and that may be millions of years old. Our approach to formulating this mission concepts utilizes automated machine learning techniques in the planning and design of space systems. We use a form of Darwinian selection to select and identify suitable number of nano-landers, the on-board instruments and control system to explore and navigate the asteroid environment. Scenarios are generated in simulation and evaluated against quantifiable mission goals such as area explored on the asteroid and amount of data recorded from the impact event. Our earlier work in this field applied to excavation robotics has shown that a machine-learning approach can discover creative solutions that exceed the capability of human devised solutions. In this work, we once again intend to compare a human-devised system to these machine evolved-systems. The results from these mission formulation and preliminary design studies will be used to identify a pathway towards a future asteroid CubeSat mission.

## 1. Introduction

The exploration of small-bodies such as asteroids and comets can give us insight into the formation of the solar-system, planetary defence and future prospect for in-situ-resource utilization. Some asteroids are thought to be ancient 'time capsules' that have remained unaltered for billions of years. Exploration of these asteroid can provide useful insight into the primordial solar system. Steady advancement in space systems technology enables us to send robotic spacecraft to asteroids and comets. Recent examples include the Hayabusa I [2] and Rosetta missions. More ambitious sample returns mission are on the way including OSIRIS-REx [1] and Hyabusa II [2]. The next technology milestone will be the ability to perform long duration surface missions on asteroids.

With rapid advancement in lightweight structural materials, miniaturization of electronics, sensors, actuators and MEMS-based instruments it has become possible to develop small, low-mass and low-cost landers for exploration of asteroid surfaces. Furthermore, with the use of Guidance, Navigation and Control (GNC) devices such as reaction-wheels and inertial measurement units it is possible for these miniature landers to perform short flights, hops and rolls to multiple locations on an asteroid.

In this paper, we study the feasibility of sending scores of nano-landers, each 1 kg in mass and volume of 1U, or 1000 $cm^3$ to an asteroid surface. These landers include science instruments such as stereo cameras, hand-lens imagers and spectrometers to characterize rock composition. Sending a network of nano-landers to an asteroid surface would give us the opportunity to perform unique and very detailed measurements of a larger spacecraft impacting an asteroid surface. The 'swarm' of nano-landers would provide multiple complementary views of the impact, the resultant seismic activity and trajectory of the ejecta.

Furthermore, this swarm can analyse the pristine, unearthed regolith that may be millions of years old and is shielded from effects of UV and cosmic rays. In-situ measurements of the artificial impact will help characterize the asteroid composition and geology. Insight into asteroid composition and impact effects can further the development of asteroid deflection techniques intended for planetary defence. A swarm of nano-landers have many advantages including working cooperatively to climb precarious slopes or vertical walls of large craters and other extreme surfaces. Multiple landers can work cooperatively by being interlinked using spring-tethers and work much like a team of mountaineers to systematically climb a slope [3,4]. In addition, multiple landers can cooperatively simplify a complex task into its components through a process of task decomposition [17, 18]. Furthermore, with a swarm, the loss of one or a few landers does not end the mission. A unique advantage to a swarm of landers is that it can operate in parallel performing many different tasks at once, hence reducing the total time required to complete time intensive tasks such as mapping or exploration.

Unlike Earth, Mars or Moon, asteroid gravity is very low which brings news opportunities and challenges. The cost of hopping or flying over an asteroid is significantly low, however high-thrust can result in a lander attaining escape velocity and tumbling in space. A conventional multi-wheeled rover would be ineffective on an asteroid surface due to the low-traction and low escape velocity. Small asteroids typically have complex gravity fields because of their highly irregular shape. This impacts surface operation, as the maximum operational speed is limited by the local gravitational field. Having a swarm of landers minimizes the risks of overshooting the maximum speed and tumbling above an asteroid. In addition, a swarm of landers can form redundant networks to effectively communicate images and videos from multiple points of view from many locations of an asteroid surface all at once. In the following sections, we present background and related work followed by description of the asteroid environment, overview of the proposed nano-lander system, methods for surface mobility and strategies for swarming followed by discussions, conclusions and future work.

## 2. Related Work

Exploration of asteroids has been a major challenge in space exploration. Much work has been done in observing asteroids by ground based telescopes and space observatories but exploring asteroid surfaces with landers is a major challenge. Several asteroid sample return missions have been launched and several others are being studied worldwide. Japan Aerospace Exploration Agency (JAXA) developed an unmanned spacecraft named Hayabusa to return a sample of material from a small near-Earth asteroid named 25143 Itokawa to Earth for further analysis. Hayabusa studied the asteroid's shape, spin, topography, colour, composition, density and history and finally landed in November 2005 and collected tiny grains of asteroid material [2]. The spacecraft also carried a 591g small rover named MINREVA (Micro/Nano Experimental Robot Vehicle for Asteroid) which unfortunately hopped off the surface of the asteroid and tumbled into space [5].

The lessons learned from the successful Hayabusa I mission led to the development of Hayabusa II asteroid sample return mission. Hayabusa II was launched in December 2014 and is expected to reach its target asteroid, 162173 Ryugu (1999 JU3) in July 2018 [2]. The Institute of Space Systems of the German Aerospace Centre (DLR) in cooperation with French space agency (CNES) built a small lander called MASCOT (Mobile Asteroid Surface Scout) to complement the sample return mission. MASCOT carries an infrared spectrometer, a magnetometer, a radiometer and a camera, and can lift off the asteroid to reposition itself for further measurement [6].

Another asteroid sample return mission is NASA's Origins, Spectral Interpretation, Resource Identification, Security, Regolith Explorer (OSIRIS-Rex) mission led by University of Arizona and was launched in September 2016. It is expected to reach its target asteroid 101955 Bennu in August 2018. The spacecraft will measure Bennu's physical, geological, and chemical properties and collect at least 60 g of regolith. Rosetta is another spacecraft built by the European Space Agency, launched in March 2004 and performed detailed study of comet 67P/Churyumov-Gerasimeko. Rosetta carried a ~98 kg lander named Philae which performed studies on elemental, isotopic, molecular and mineralogical composition of the comet, characterized the physical properties of the surface and subsurface material and the magnetic and plasma environment of the nucleus [1].

NASA JPL also developed a planetary mobility platform called "spacecraft/rover hybrid" that relies on internal actuation. With the help of three mutually orthogonal flywheels and external spikes, the platform can perform both long excursions by hopping and short, precise traverses through controlled tumbles [7,8]. Another autonomous microscale surface lander developed is PANIC (Pico Autonomous Near-Earth Asteroid In-Situ Characterizer). PANIC has a shape of a regular tetrahedron with an edge length of 35 cm, mass of 12 kg and utilizes hopping as a locomotive mechanism in microgravity [9].

In our work, we look towards the next milestone in asteroid exploration, namely to perform long-duration surface exploration and analysis. Miniaturization of space electronics makes it viable to build miniature landers. However, to attain the full benefits of utilizing miniature landers require sending many of them that can operate independently and cooperate in teams.

### 3. The Asteroid Environment

Most asteroids are irregular in shape and are pock-marked by impact craters. An asteroid's shape can be described using a polyhedron, whose surface consists of a series of triangles. In this paper, we use an irregular polyhedron model of asteroid Castalia (Fig. 1) [10].

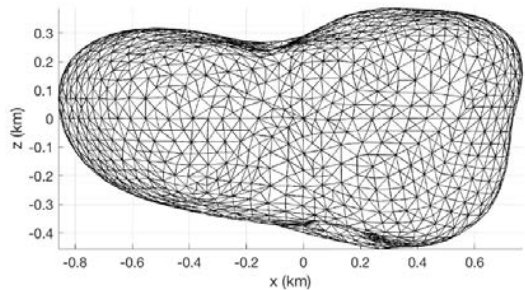

Fig. 1. Polyhedron model of asteroid Castalia.

Using the polyhedron model, the gravitational surface potential of an asteroid can be determined [11, 12]. Fig. 2 shows the distribution of acceleration due to gravity in m/s$^2$ in the x-z plane for the polyhedron model of asteroid shown in Fig. 1 with a density of 2.1 g/cm$^3$.

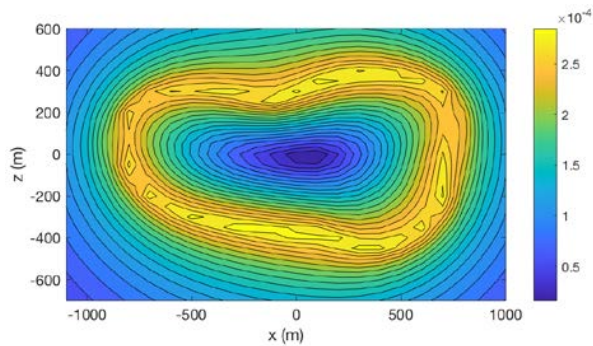

Fig. 2. Acceleration due to gravity for the polyhedron model of asteroid Castalia in the x-z plane.

We use this representative gravitational model of asteroid Castalia to perform our design and control studies described in the paper.

### 4. Nano-Lander System Overview

Each nano-lander has a mass of 1 kg and a volume of 1U or 1000 cc. Fig. 3 shows the internal and external views of the lander. Three mutually orthogonal reaction wheels are mounted on adjacent faces of the lander to maximize its moment of inertia and allow more space for scientific payload and avionics. The system consists of one spike on each corner to facilitate hopping and tumbling.

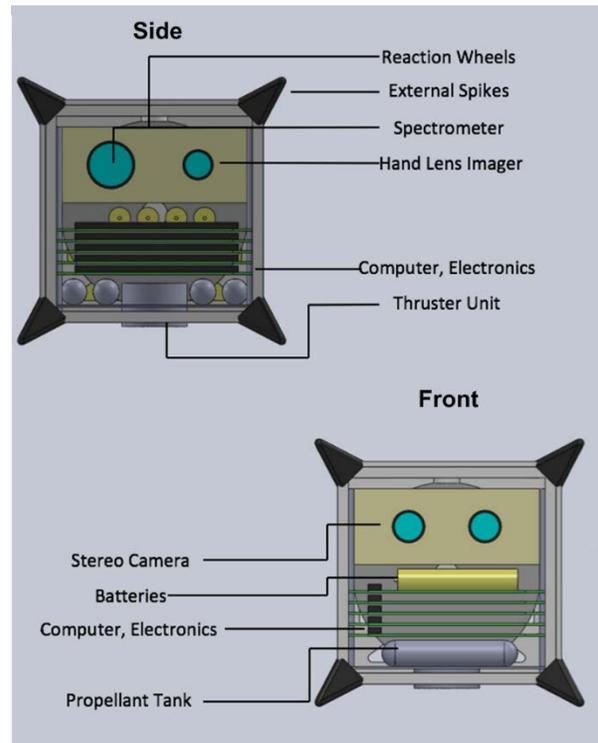

Fig. 3. CAD Design of a 1U Nano-lander Concept.

The top half of the proposed nano-lander consists of a scientific payload including a miniature x-ray spectrometer, hand-lens imager and stereo cameras. Below the instruments are the avionics which consists of an on-board computer, IMU, radio transceivers, power board and batteries. The lower-half consists of the propulsion system, with storage tanks for the fuel and the oxidizer connected to a min-thruster. Solar cells with an anti-dust cohesion coating are placed on the external surfaces of the lander and slot antennas are located around the cells.

*4.1 Communication*

Each nano-lander is required to communicate with its neighbours and to the mothership in orbit around the asteroid. Since each lander hops and rolls around the asteroid surface, use of a deployable antenna is not feasible. A feasible alternative is to place multiple patch antennas on each face of the lander. For our design, we consider use of slot antennas that are complementary dipole antennad and have roughly omnidirectional radiation patterns. Its undesirable to radiate into the electronics and hence the antennas have a reflector backing to radiate outwards from the spacecraft.

Fig. 4 and 5 show the radiation pattern of the reflector backed slot antenna. The slot has a dimension of 75 × 2 mm fed with a 50 Ω lumped port at the centre.

The reflector has a dimension of 10 × 10 cm with 0.7 cm spacing. The antenna has a forward gain of 10.2 dBi at 3.8GHz.

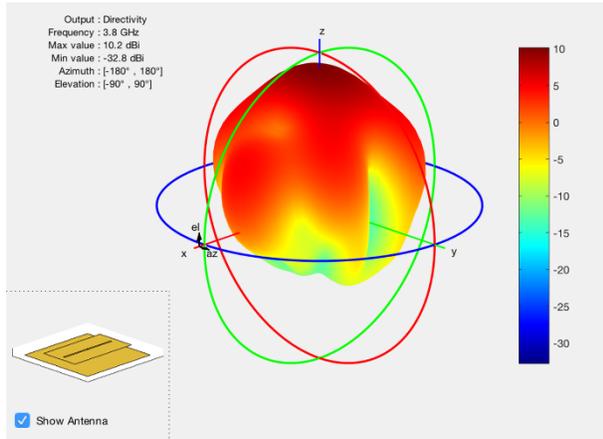

Fig. 4. Radiation pattern of the reflector backed slot antenna.

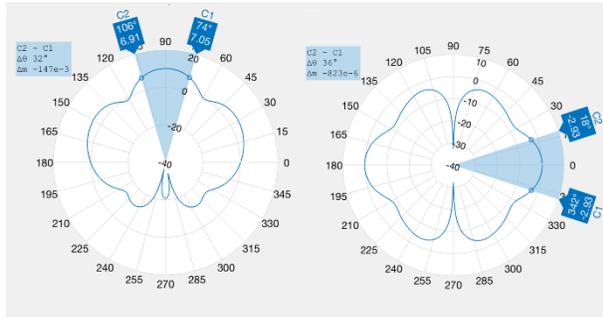

Fig. 5. (a) $0^0$ Azimuth beam-width. (b) $0^0$ Elevation beam-width.

The $0^0$ azimuth beam-width is $32^0$ and the $0^0$ elevation beam-width is $36^0$. Fig. 6 shows the return loss of the antenna.

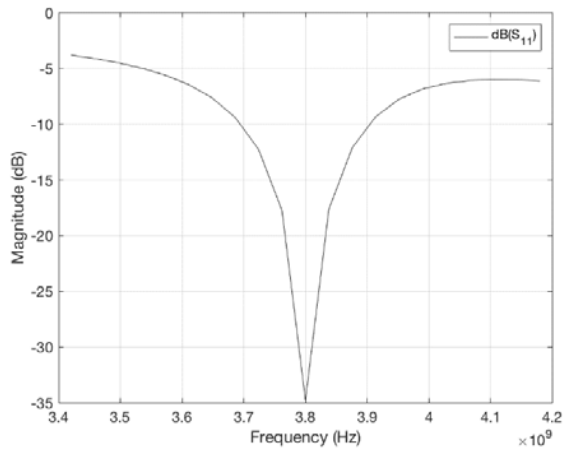

Fig. 6. Return loss of the reflector backed slot antenna.

## 5. System Mobility

*5.1 Mobility Utilizing Propulsion*

Propulsion enables each lander to perform ballistic hops on the asteroid surface. This section describes the control approach of each lander to perform ballistic hops using the propulsion unit or on-board 3-axis reaction wheel system [13]. During each ballistic hop, the thrust generated by the propulsion unit acts along $+z$ axis with gravity acting along $-z$ axis. The 3-axis reaction wheel system applies torque about the lander's three principle axes to change its Euler angles and angular velocities according to a PD control law shown as below:

$$\tau_{rw} = -K_p(e_{des} - e_{act}) - K_d(\omega_{des} - \omega_{act}) \quad (1)$$

Using RP-1 as the fuel, $H_2O_2$ as the oxidizer, and with a throat diameter of 60 μm, the propulsion unit can deliver a thrust of 0.0445 N with an $I_{sp}$ of 370 s. Fig. 7 shows a ballistic hop on an asteroid gravity of 0.001 m/s$^2$.

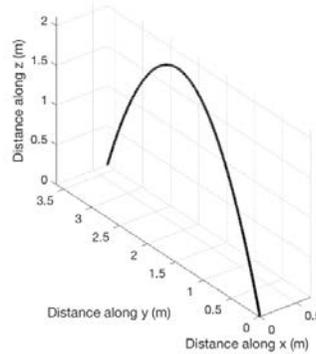

Fig. 7. Trajectory of a nano-lander performing a rocket-propelled ballistic hop on an asteroid.

The lander can hop 3.8 m by burning only 20 mg of RP1-$H_2O_2$ propellant. Fig. 8 shows the velocity of the lander during its flight. The maximum velocity achieved is 0.07 m/s which is lower than the escape velocity of the asteroid considered.

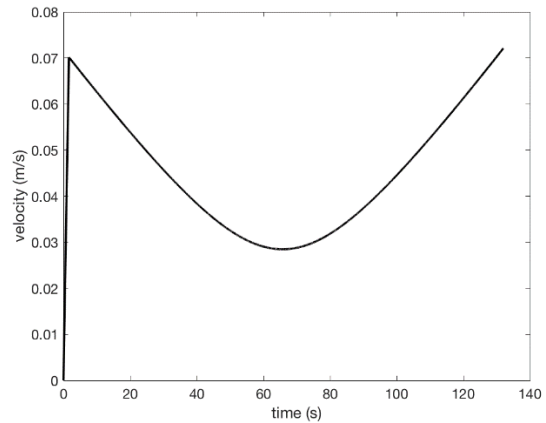

Fig. 8. Velocity of a lander during the ballistic hop.

Each reaction wheel has a mass of 100 g, radius of 4.3 cm and can produce a maximum torque of 10 mN. Each reaction wheel operates according to the PD law and provides a torque to change the orientation of the lander. Fig. 9 and 10 shows the Euler angles and angular velocity respectively of a lander during each ballistic hop. The reaction wheel simplifies the control architecture by maintaining a steady, commanded direction while hopping.

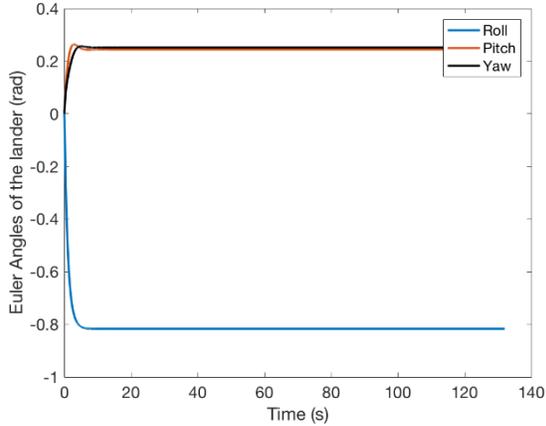

Fig. 9. Euler angles of the lander.

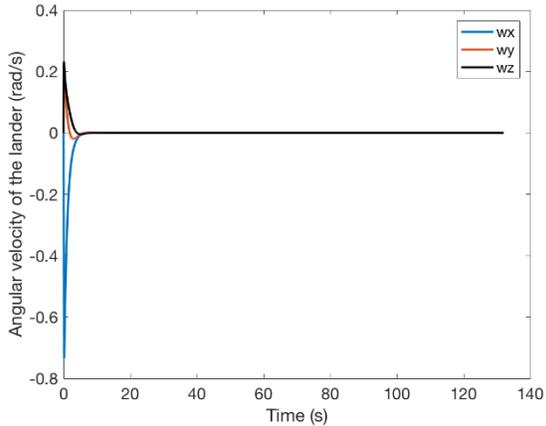

Fig. 10. Angular velocity of the lander during ballistic hop.

*5.2 Mobility Utilizing Reaction Wheels*

With the reaction wheels, the nano-lander system is capable of two-modes of mobility: tumbling and hopping. The lander can produce momentarily large reaction forces at the surface by exerting internal torques with the help of the 3-reaction wheels. The contact forces between the spikes and the surface can be modelled as a spring-damper force normal to the surface, and a Coulomb friction force tangential to the surface [14]. The spikes act as a pin-joint and the model can be described by the angle, $\theta$, and torque, $\tau$, as can be seen in Fig. 11 along with the other parameters [7,8].

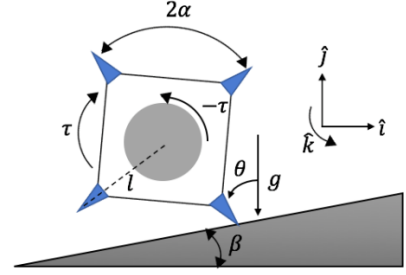

Fig. 11. 2D model of the lander with all physical parameters.

In the tumbling mode, the system pivots to the right and lands on the next consecutive spike. The hopping mode consists of a stride phase, when the system is supported by a single spike, and a flight phase when the spike leaves the ground (negative normal force). The equations of motion for the stride phase can be written as:

$$\ddot{\theta}(t) = \frac{m_s g l \sin \theta(t) - \tau(t)}{I_s + m_s l^2} \quad (2)$$

For tumbling, the minimum torque required to initiate angular acceleration ($-\ddot{\theta}$) from rest is given by (9). The corresponding minimum reaction wheel angular velocity required is given as follows:

$$\tau_{min} = m_s g l \sin(\alpha + \beta) \quad (3)$$

For hopping, the applied torque must be greater than that for tumbling $\tau \gg m_s g l \sin(\theta)$ and the corresponding reaction wheel velocity required to hop a distance $d_h$ is given by:

$$\omega_r = \sqrt{2 m_s g l (1 - \cos(\alpha + \beta))/(\eta I_r)} \quad (4)$$

Both the hopping angle and the lateral distance covered are a function of $\tau$ and $\omega_r$ as shown by equation (5), (6) and (7):

$$\omega_r(d_h) = \sqrt{\frac{d_h g}{\eta^2 l^2 \sin(2(\alpha + \beta))}} \quad (5)$$

$$\theta_h = \alpha - \frac{\eta I_r \omega_r^2}{2\tau} \quad (6)$$

$$d_h = \sin(2\alpha - \eta I_r \omega_r^2 / \tau)(\eta l \omega_r)^2 / g \quad (7)$$

For both hopping and tumbling, the torque required is produced using a hybrid control algorithm, where the reaction wheels are slowly accelerated to the desired angular velocity and then impulsively braked. This control approach also restricts the reaction wheel to reach its saturation speed. Fig. 12 shows the torque required for tumbling at various $g$ and Fig. 13 shows the torque required for hopping 1 m, 5 m and 10 m. The length of the spike is 0.1 m for the simulations. Fig. 14 shows the hopping trajectories of each lander for different values of reaction wheel angular velocities and torque.

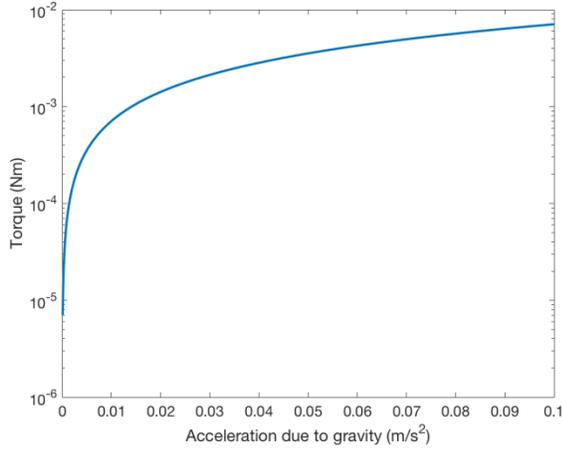

Fig. 12. Torques for tumbling motion (spike = 0.1 m).

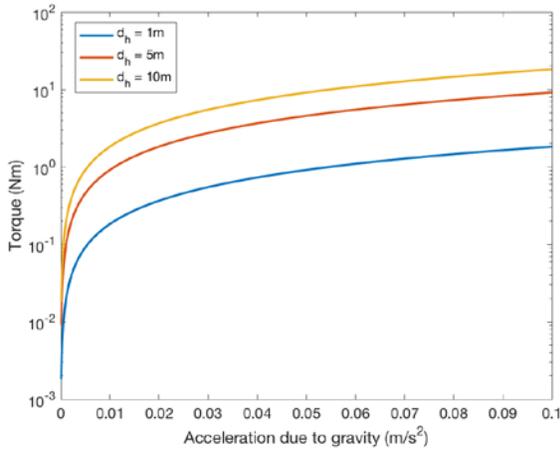

Fig. 13. Torques for hopping motion (spike = 0.1m).

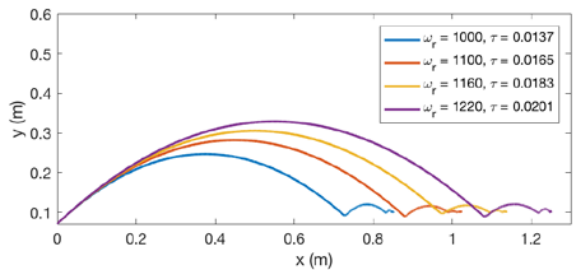

Fig. 14. Hopping trajectories of the lander for different values of $\omega_r$ and $\tau$ ($\omega_r$ is in rpm, $\tau$ is in Nm, $g$ = 0.001 m/s² and $l$ = 0.1 m)

## 6. Maximum Coverage Algorithm for Multiple Landers

Utilizing the mobility strategies described earlier, we can enable a swarm of nano-landers to achieve mobility on the asteroid surface and perform exploration. The asteroid surface is scattered with fragments of rocks and large boulders. These objects maybe dangerous obstacles for the nano-landers. A key requirement is to avoid them. So, the system of multiple nano-landers deployed on the asteroid surface are required to avoid obstacles, while maximizing area coverage. A third requirement is that the nano-landers maintain multiple communication links so that acquired science data maybe communicated effectively to a mothership.

In this section, we describe an algorithm developed to distribute a fleet of $N$ nano-landers on an asteroid surface (Table 1). We use the concept of virtual forces to repel each lander from the rest of the fleet.

Table 1. Pseudo-code for area coverage maximization using a fleet of lander.

**Algorithm**: Maximize coverage for multiple landers

**Require**: Initial position, orientation for all landers $i = 1$ to $N$;
1. Compute the Euclidean distance between each lander;
2. Compute the degree $D$ for each lander based on the communication range ($R_c$);
3. Compute the Euclidean distance between each lander and its neighboring obstacle;
4. **for** $k = 0$ to $K$ **do**
5.    **for** $i = 1$ to $N$ **do**
6.      Compute the net force on lander $i$, according to (8) - (11);
7.    **end for**
8.    **for** $i = 1$ to $N$ **do**
9.      **for** $t = 0$ to $k+1$ **do**
       Move each lander $i$ according to (12)
     **end for**
10.    At $t = k+1$, compute the new
11.    Euclidean distances and degree $D$;
   **end for**
12. **end for**

The nano-landers are all identical and operate in a distributed fashion without relying on a single surface asset. They have equal sensing range ($R_s$) and equal communication range ($R_c$). Each lander can communicate its location and orientation to its neighbours and has a laser rangefinder to locate and characterize obstacles.

In our area coverage algorithm, the landers interact with each other through a combination of global repulsion combined with local, limited attraction. The repulsion and attraction are achieved using a concept called *virtual forces* that we simulate to enable collective control over the nano-landers. The modelled *virtual forces* used to position the landers are of three kinds: $F_{cov}$, $F_{com}$ and $F_{obs}$. $F_{cov}$ causes the landers to repel each other to maximize the sensing range of the target area, $F_{coms}$ constrains the degree of communication links for each lander by attracting landers (locally) when they are on the verge of losing connection. $F_{obs}$ causes the landers to move away from neighboring obstacles [15]. Considering a network of $N$ landers $1, 2, 3... N$ with positions $r_1, r_2, r... r_N$ respectively and $\|r_{ij}\|$ representing the Euclidean distance between landers $i$ and $j$, $F_{cov}$ and $F_{coms}$ are defined in (8) and (9) respectively:

$$F_{cov}(i,j) = \left(\frac{C_{cov}}{\|r_{ij}\|}\right)\left(\frac{r_i - r_j}{\|r_{ij}\|}\right) \quad (8)$$

$$F_{com}(i,j) = \begin{cases} (-C_{com}\|r_{ij}\|)\left(\frac{r_i - r_j}{\|r_{ij}\|}\right) & \text{if degree} < D \\ 0 & \text{otherwise} \end{cases} \quad (9)$$

Similarly, for $L$ obstacles $1, 2, 3... L$ with positions $r_1, r_2, r_3... r_L$ respectively and $\|r_{il}\|$ representing the Euclidean distance between lander $i$ and obstacle $l$, $F_{obs}$ is defined as follows.

$$F_{obs}(i,l) = \left(\frac{C_{obs}}{\|r_{il}\|}\right)\left(\frac{r_i - r_l}{\|r_{il}\|}\right) \quad (10)$$

Where, $C_{cov}$, $C_{com}$ and $C_{obs}$ are the force constants and the net force experienced by lander $i$ can be expressed as follows:

$$F(i) = \sum_{j=1, j\neq i}^{N}(F_{cov}(i,j) + F_{com}(i,j)) + \sum_{k=1}^{L} F_{obs}(i,k) \quad (11)$$

The equation of motion for lander $i$ can then be formulated as:

$$m_i \frac{d^2 r_i}{dt^2} + \mu_i \frac{dr_i}{dt} = F(i) \quad (12)$$

Where, $m_i$ is the mass and $\mu_i$ is the damping factor of lander $i$. When the distance between two landers tends to zero, $\|F_{cov}\| \to \infty$ to avoid collisions. When the degrees of connection between a lander and neighbor is less than $D$, $\|F_{com}\| > 0$ to prevent loss of connection. Similarly, $\|F_{obs}\| \to \infty$ when the distance between a lander and an obstacle tends to zero to avoid collisions.

For simulation of the stated algorithm, we considered 40 landers deployed at random positions inside a square test area. Each lander has a communication range, $R_c$ = 5 units and sensor range, $R_s$ = 2.5 units. The target area consists of obstacles of random sizes at random positions. The 40 landers must move in the 2-D space in such a way that it maximizes the coverage area, avoiding collision with each other and the obstacles and maintaining a degree of communication links, $D$ = 3. Fig. 15 shows the lander positions at different times. The landers disperse to maximize distance while maintaining a communication link between two neighbours. The red dots are the obstacles, black dots the landers and the lines connecting them are the active communication links

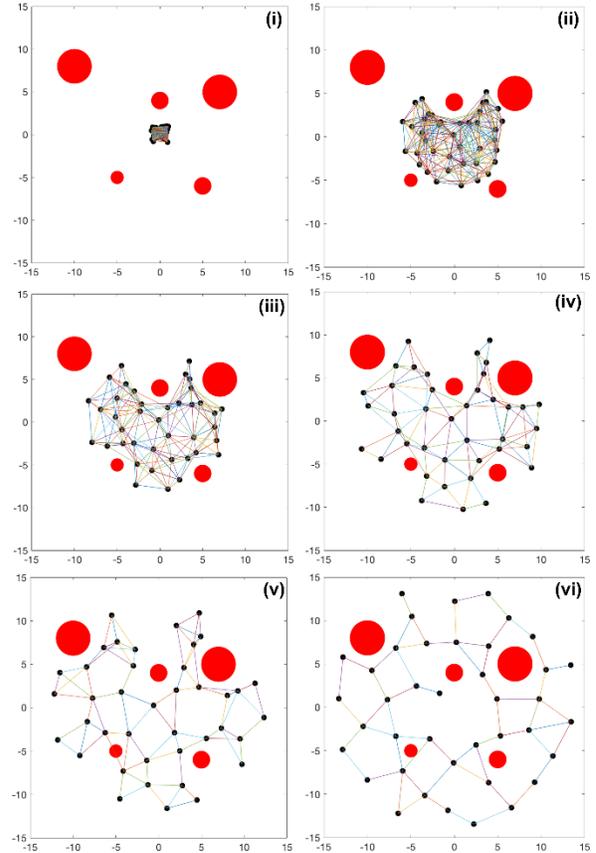

Fig. 15. Simulation of a system of 40 landers at timestep 0, 15, 30, 60, 100 and 200.

Fig. 16 shows the variation of the coverage area with time for different values of $D$ = 2,3,4,5, and 6. The swarm of nano-landers can provide unique and very detailed measurements of a spacecraft impacting onto the asteroid surface. Fig. 17 shows a second simulation of a swarm of robots being simulated to repel a target area and form 'donut' around the area. This will enable the swarm to track and record the impact event and collect data from multiple viewpoints. The red dots are the obstacles and the black dots are the landers. The landers were placed randomly on the target area and the impact event is supposed to take place at coordinates (3, -1). Each lander positions itself to be at a safe distance from the target impact site, while avoiding obstacles.

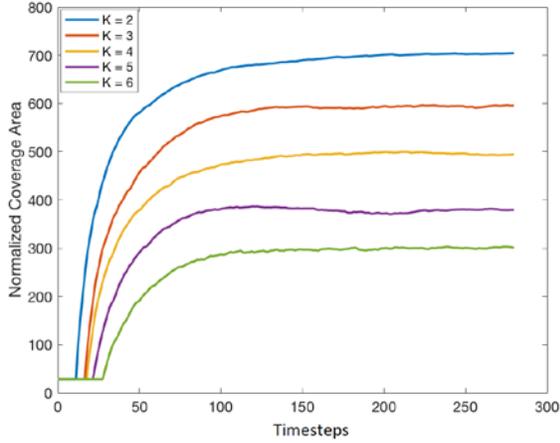

Fig. 16. Area coverage by a swarm of 40 robots with respect to settling timesteps.

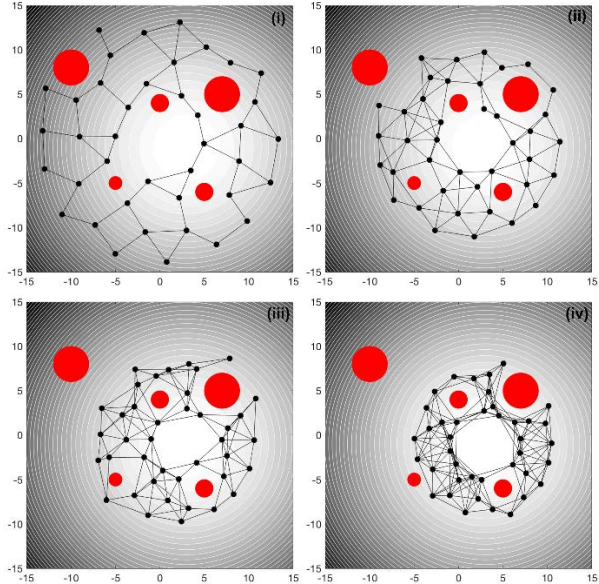

Fig. 17. Simulation of a system of 40 landers commanded to avoid a target impact site at 0, 50, 100 and 150 timesteps.

## 7. System Optimization

In the previous section, we have shown that a swarm of nano-landers have the potential to solve the intended task. However, what is apparent is that there are several system parameters that a critical for enabling area coverage maximization. Selection of swarm parameters such as the number of landers, $N$ and $C_{cov}$, $C_{com}$ and $C_{obs}$ are non-intuitive. This is a scenario where Evolutionary Algorithms have potential, as they can generate good enough solutions through a directed, trial and error search.

Here we use Evolutionary Algorithms to find the minimum number of landers required for maximum coverage of a target area with redundant communication links. The genotype of the EA population (see Table 2) are represented by binary numbers so that they can be easily manipulated by standard genetic operators such as crossover and mutation. Each individual is represented by 19 bits with the first seven bits representing the number of landers, next three bits representing the degree, next four bits representing the force constant $C_{cov}$ and the final five bits representing the force constant $C_{com}$ as shown in Table 2. Furthermore, we assume $C_{cov} = C_{obs}$.

Table 2: Genotype of the nano-lander swarm.

| N | | | | | | | D | | | $C_{cov}$ | | | | $C_{com}$ | | | | |
|---|---|---|---|---|---|---|---|---|---|---|---|---|---|---|---|---|---|---|
| a | b | c | d | e | f | g | h | i | j | k | l | m | n | o | p | q | r | s |
| Binary [1,0] | | | | | | | | | | | | | | | | | | |

### 7.1 Fitness Function

The total coverage area is estimated by considering the formation of the swarm as a non-self-intersecting polygon with the coordinates of the landers $(x_1, y_1) \ldots (x_N, y_N)$ as the vertices. The area can then be calculated as:

$$A = \left\| \frac{1}{2} \left( \begin{vmatrix} x_1 & x_2 \\ y_1 & y_2 \end{vmatrix} + \begin{vmatrix} x_2 & x_3 \\ y_2 & y_3 \end{vmatrix} + \cdots + \begin{vmatrix} x_N & x_1 \\ y_N & y_1 \end{vmatrix} \right) \right\| \quad (13)$$

The normalized fitness of the area function can then be calculated as

$$A_n = \frac{A}{900} \quad (14)$$

The degree of communication links for each lander is calculated by calculating the number of landers within a distance of 5 units and then averaged for all the landers. The normalized fitness is then calculated as

$$D_n = \frac{D_{ach}}{D_{req}} \quad (15)$$

where, $D_{ach}$ is the average degree achieved and $D_{req}$ is the average degree required. The settling time objective function is then calculated by determining the time step at which the swarm settles down and doesn't move further (16). Similarly, the energy objective function is calculated by determining the number of hops each lander takes to until the formation is settled (17). To calculate the normalized fitness of the settling time objective function and energy objective function, first we take the time required and energy dissipated by a 40-lander swarm to cover a target area of 30×30 units. Then we specify the normalized fitness relative to the time and energy taken by the 40-lander system as shown below:

$$T_n = \frac{t_{40} - \Delta_t}{t_{40}} \quad (16)$$

$$E_n = \frac{e_{40} - \Delta_e}{e_{40}} \quad (17)$$

where, $t_{40}$ and $e_{40}$ are reference values for time and energy corresponding to the 40-lander system, $\Delta_t$ and $\Delta_e$ are the differences in the actual values and reference values. The overall fitness of the system is then determined by taking the weighted average as:

$$F = 0.5A_n + 0.25D_n + 0.125T_n + 0.125E_n \quad (18)$$

An elitist non-dominated sorting algorithm (NSGA-II) is used in this paper [16]. The initial parent population, $P_t$, is created randomly of size $M$ which is then sorted based on the non-domination and then assigned a rank based on the fitness which is equal to its non-dominant level. The initial population then undergoes crossover and mutation to produce the set of offspring population $O_t$ of size $M$. Both the parents and children are then combined to produce $C_t = P_t \cup O_t$ of size $2M$. The population $C_t$ is then sorted via non-dominance and assigned a rank. The first $M$ individuals of the set $C_t$ based on the non-dominant level is then selected for the next generation. The next population $P_{t+1}$ of size $M$ then again undergo selection, crossover and mutation. The process is repeated until the system achieves the desired fitness.

For our analysis, we have considered an initial population of 50 with a crossover probability of 0.8 and a mutation probability of 0.2. For non-dominated sorting, we have considered four fitness objectives. The first fitness is the coverage area, second fitness is the average communication links (degree) of each robot, third fitness is the time step required for maximum coverage and the fourth fitness is the average energy consumed by each lander. Each of these fitness values are then normalized between 0 and 1, with 1 representing the fittest value. The weighted average of the four fitness values are then multiplied with their corresponding weights 0.5, 0.25, 0.125 and 0.125 respectively which represents the overall fitness of the system (18). Moreover, to test the redundancy of the communication links, 10% of the landers are killed off in every generation. Each lander has a communication range, $R_c$, of 5 units and the goal is to cover a target area of $30 \times 30$ units.

Fig. 18 shows the fitness of each objective function over 40 generations. Fig. 19 shows the overall fitness of the swarm over 40 generations.

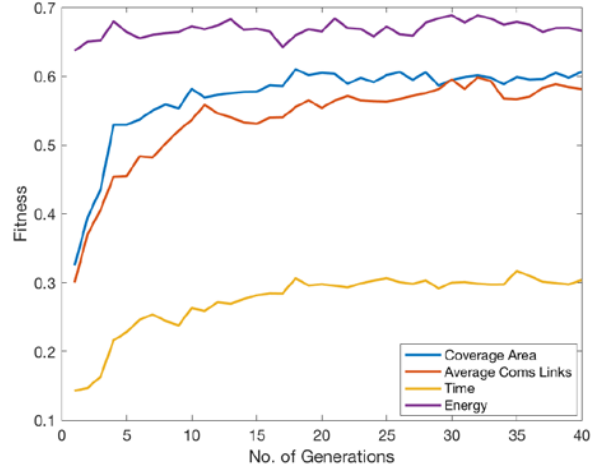

Fig. 18. Component fitness of a swarm evolved over 40 generations. Results obtained from an average of 5 Evolutionary Algorithm runs.

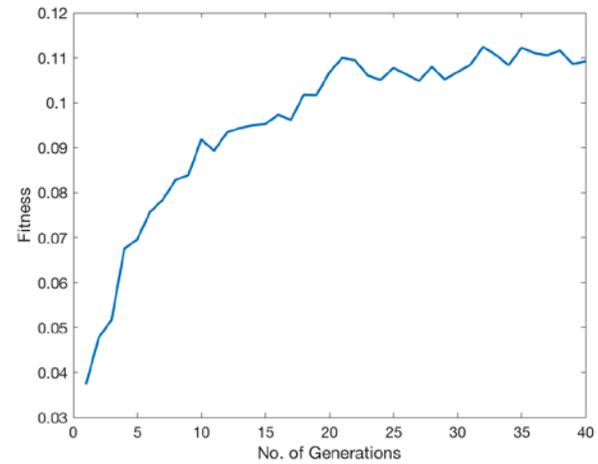

Fig. 19. Overall fitness of a swarm evolved over 40 generations. Results obtained from an average of 5 Evolutionary Algorithm runs.

## 8. Discussion

Our approach to designing and utilizing a swarm of nano-landers to perform asteroid exploration presents important challenges but also new opportunities. We show the initial feasibility of a decentralized swarm of robots that can be used to attain maximum area coverage over an asteroid surface while avoiding obstacles and maintaining sufficient number of redundant communication links. In addition, the swarm can be used to position themselves around a target event and obtain multiple views at once.

This initial analysis shows the potential of a swarm, but the parameters required to obtain suitable performance was not well understood. Further, it was unclear if the system could be further optimized. This is where machine learning approaches such as Evolutionary Algorithms shine. Where there is limited domain knowledge of the task at hand, the algorithm can find desirable solutions through a process of directed trial and error search.

The results show that a system of 65 landers with degree D = 5 as the best solution (found to date) for maximum coverage area of 30 × 30 units. Using EAs we have effectively improved the overall system fitness, including maximum area covered and increased number of communication links, while maintaining a nearly constant energy consumption. This shows the algorithm is effective in improving energy utilization of the swarm. With a more dispersed system, the settling time inevitably increases. Overall, our approach shows a promising pathway towards further refining of our system design parameters towards detailed design of the mission concept.

## 8. Conclusions

The next major phase of asteroid exploration will require sending landers to perform surface exploration. We have analysed the preliminary feasibility of operating scores of nano-landers, each 1 kg in mass and volume of 1U, or 1000 cm$^3$ on an asteroid surface. These landers would hop, roll and fly over the asteroid surface. The landers would include science instruments such as stereo cameras, hand-lens imagers and spectrometers to characterize rock composition. A network of nano-landers situated on the surface of an asteroid can provide unique and detailed in-situ measurements of a spacecraft impacting onto an asteroid surface. In this work, we demonstrate an algorithm that utilizes the concept of *virtual forces* to enable a decentralized swarm of nano-landers to effectively attain maximum area coverage for exploration and to position themselves to witness an impact event from multiple viewpoints. Our approach model multibody dynamical systems and uses Evolutionary Algorithms to further optimize for area coverage and communication performance. The results show a promising pathway towards field study in a more detailed, simulated asteroid surface environment.

## Nomenclature

$K_p$, $K_d$ = Proportional and derivative controller gains
$e_{des}$, $e_{act}$ = Desired and actual Euler angles
$\omega_{des}$ and $\omega_{act}$ = Desired and actual angular velocity
$m_s$ = Mass of the lander
$g$ = Acceleration due to gravity
$l$ = Length of spike
$\tau$ = Applied torque by reaction wheel
$I_s$, $I_r$ = Moment of inertia of lander and reaction wheel
$\eta$ = Energy transfer efficiency